\documentclass[10pt,twocolumn,letterpaper]{article}

\usepackage{cvpr}              

\usepackage{graphicx}
\usepackage{amsmath}
\usepackage{amssymb}
\usepackage{booktabs}
\usepackage{multirow}
\usepackage{xcolor}
\usepackage{graphicx}
\usepackage{makecell}
\usepackage{balance}
\usepackage[symbol]{footmisc}
\usepackage[pagebackref=true,breaklinks=true,letterpaper=true,colorlinks,bookmarks=false,citecolor=blue,linkcolor=blue]{hyperref}

\usepackage[capitalize]{cleveref}
\crefname{section}{Sec.}{Secs.}
\Crefname{section}{Section}{Sections}
\Crefname{table}{Table}{Tables}
\crefname{table}{Tab.}{Tabs.}

\def\cvprPaperID{10140} 
\def\confName{CVPR}
\def\confYear{2023}

\newcommand{\latentshift}{Latent-Shift\xspace}

\begin{document}

\title{Latent-Shift: Latent Diffusion with Temporal Shift \\for Efficient Text-to-Video Generation}

\author{Jie An$^{1,2}$\footnotemark[1] \quad Songyang Zhang$^{1,2}$\footnotemark[1] \quad Harry Yang$^2$ \quad Sonal Gupta$^2$ \quad Jia-Bin Huang$^{2,3}$ \\ Jiebo Luo$^{1,2}$ \quad Xi Yin$^2$\\
$^1$University of Rochester \quad
$^2$Meta AI \quad
$^3$University of Maryland, College Park\\
\tt\small\{jan6,jluo\}@cs.rochester.edu \quad szhang83@ur.rochester.edu \\ \tt\small\{harryyang,sonalgupta,yinxi\}@meta.com \quad jbhuang@umd.edu}


\twocolumn[{
\renewcommand\twocolumn[1][]{#1}
\maketitle
\begin{center}
    \centering
    \includegraphics[width=0.97\textwidth]{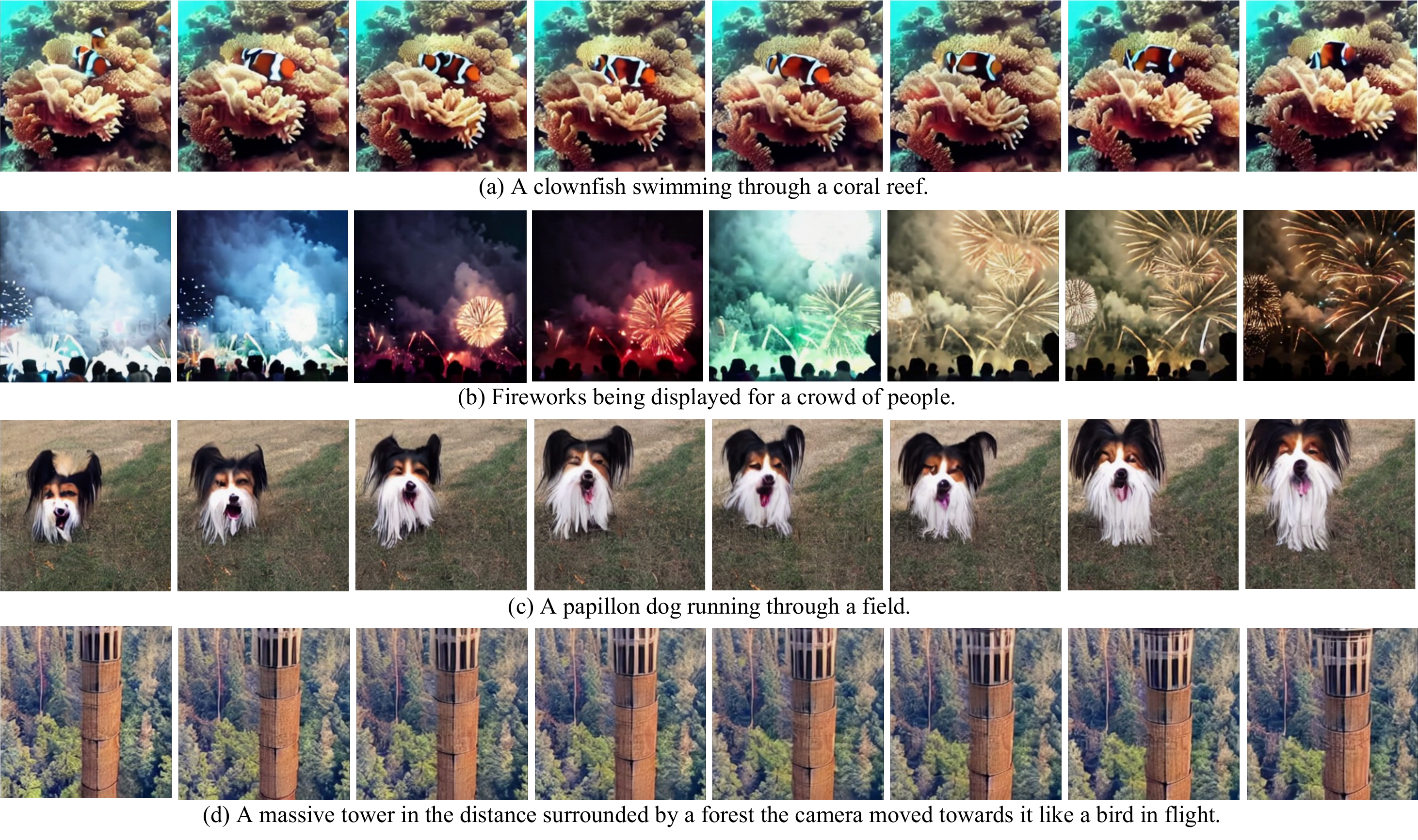}
    \vspace{-3mm}
     \captionof{figure}{Text-to-Video generation results. Our method can generate rich contents with meaningful motions including both object/scene motions ((a-c)) and camera motion ((d)). please check our project page \url{https://latent-shift.github.io} for video samples.}
    \label{fig:teaser}
\end{center}
}]

\footnotetext[1]{Equal contribution, ordered alphabetically. This work was done when Jie and Songyang interned at Meta AI.}

\begin{abstract}
We propose Latent-Shift --- an efficient text-to-video generation method based on a pretrained text-to-image generation model that consists of an autoencoder and a U-Net diffusion model. 
Learning a video diffusion model in the latent space is much more efficient than in the pixel space. 
The latter is often limited to first generating a low-resolution video followed by a sequence of frame interpolation and super-resolution models, which makes the entire pipeline very complex and computationally expensive. 
To extend a U-Net from image generation to video generation, prior work proposes to add additional modules like $1$D temporal convolution and/or temporal attention layers. 
In contrast, we propose a parameter-free temporal shift module that can leverage the spatial U-Net as is for video generation. 
We achieve this by shifting two portions of the feature map channels forward and backward along the temporal dimension. 
The shifted features of the current frame thus receive the features from the previous and the subsequent frames, enabling motion learning without additional parameters.
We show that~\latentshift achieves comparable or better results while being significantly more efficient. Moreover, \latentshift can generate images despite being finetuned for T2V generation.

\end{abstract}

\section{Introduction}
\label{sec:introduction}
In the last two years, tremendous progress has been made in generative modeling in AI. Text-to-Image (T2I) generation systems~\cite{dalle2,imagen,yu2022scaling,oran2022make,mou2023t2i} trained on large-scale text-image pairs can now generate high-quality images with novel scene compositions.
In particular, latent diffusion models for T2I generation have garnered high interest because of the efficient modeling with fewer modules.

Recent work extends T2I models for text-to-video (T2V) generation. The two main challenges in T2V generation are the lack of high-quality text-video data at scale and the complexity of modeling the temporal dimension. There are two mainstream frameworks: 1) Transformer with Variational Auto Encoders (VAE); 2) diffusion models with U-Net. CogVideo~\cite{hong2022cogvideo} and Phenaki~\cite{villegas2023phenaki} are based on VAE and Transformer to learn T2V generation in the latent space. Make-A-Video~\cite{singer2023makeavideo} and Imagen Video~\cite{Ho2022ImagenVH} are based on diffusion models to learn video generation in the pixel space and have shown better performance than their Transformer+VAE counterparts. However, due to the complexity of video modeling, 
pixel-based T2V diffusion models must compromise to generate a low-resolution video first ($64\times64$ in Make-A-Video and $40\times24$ in Imagen Video), followed by a sequence of super-resolution and frame interpolation models (see Tab.~\ref{tab:model_components} for details). This makes the entire pipeline complicated and computationally expensive. 

A generative AI system's efficiency is essential because it impacts the user experience when interacting with these tools. Additionally, a simpler model architecture aids further research and development on top of it.
In this paper, we propose~\latentshift, which is an efficient model that can generate a two seconds video clip with $256\times256$ resolution without additional super-resolution or frame interpolation models. 

Our work builds on the T2I latent diffusion model.
When expanding the U-Net from T2I to T2V generation, we carefully choose not to increase the model complexity. Unlike prior work that adds additional temporal convolutional layers~\cite{singer2023makeavideo} and/or temporal attention layers~\cite{singer2023makeavideo,ho2022video,zhou2022magicvideo} 
to expand the U-Net for temporal modeling, we use a parameter-free temporal shift module as motivated from~\cite{lin2019tsm,munoz2021temporal}. During training, we shift a few channels of the spatial U-Net feature maps forward and backward along the temporal dimension. This allows the shifted features of the current frame to observe the features from the previous and the subsequent frames and thus help to learn temporal coherence. We show in our experiments that~\latentshift
achieves better performance than latent video diffusion models with temporal attention
while having fewer parameters and thus being more efficient.  

In summary, our main contributions are three-fold: 
\begin{itemize}
    \item We propose a novel temporal shift module to leverage a T2I model as-is for T2V generation without adding any new parameters.
    \item We show that our~\latentshift model finetuned for video generation can also be used for T2I generation, which is a unique capability of the parameter-free temporal shift module.
    \item We demonstrate the effectiveness and efficiency of \latentshift through extensive evaluations on MSR-VTT, UCF-101, and a user study. 
\end{itemize}

\section{Related Work}
\label{sec:related_work}

\noindent\textbf{Text-to-Image Generation.} 
Early work in T2I generation~\cite{xu2018attngan,zhang2017stackgan} are focused on GAN-based~\cite{nilsback2008automated} extensions to generate images in simple domains like flowers~\cite{nilsback2008automated}, birds~\cite{welinder2010caltech}, {\it{etc}}. Recent work leverage better modeling techniques like Transformer with VAE or diffusion models to enable zero-shot T2I generation with compelling results. For example, CogView~\cite{ding2021cogview, girish2020understanding}, DALLE~\cite{ramesh2021zero}, and Parti~\cite{yu2022scaling} train an auto-regressive Transformer on large-scale text-image pairs for T2I generation. Make-A-Scene~\cite{oran2022make} additionally adds a scene control to allow more creative expression. On the other hand, GLIDE~\cite{nichol2021glide}, DALLE2~\cite{dalle2}, and Imagen~\cite{imagen} leverage diffusion models and achieve impressive image generation results. These diffusion-based models are trained on the pixel space and require additionally trained super-resolution models to achieve a high resolution. Latent diffusion can generate high-resolution images directly by learning a diffusion model in the latent space to reduce the computational cost. We extend the latent diffusion model for T2V generation. 

\noindent\textbf{Text-to-Video Generation.}
Similar to the evolution in T2I generation, early T2V generation 
 methods~\cite{mittal2017sync,pan2017create,li2018video} are based on GAN and applied to constrained domains like moving digits or simple human actions. Due to the challenges in modeling video data and a need for large-scale, high-quality text-video datasets, the priors of T2I in both modeling and data are leveraged for T2V generation. For example, N\"UWA~\cite{wu2022nuwa} formulates a unified representation space for image and video to conduct multitask learning for T2I and T2V generation. CogVideo~\cite{hong2022cogvideo} adds temporal attention layers to the pretrained and frozen CogView2~\cite{ding2022cogview2} to learn the motion. Make-A-Video~\cite{singer2023makeavideo} proposes to finetune from a pretrained DALLE2~\cite{dalle2} to learn the motion from video data alone, enabling T2V generation without training on text-video pairs. Video Diffusion Models~\cite{ho2022video} and Imagen Video~\cite{Ho2022ImagenVH} perform joint text-image and text-video training by considering images as independent frames and disabling the temporal layers in the U-Net. Phenaki~\cite{villegas2023phenaki} also conducts joint T2I and T2V training in the Transformer model by considering an image as a frozen video.
 
While the advance in video generation is exciting, the entire pipeline for video generation can be very complex. As shown in Tab.~\ref{tab:model_components}, Make-A-Video has $6$ models to generate a high-resolution video, and Imagen Video has $8$ models, as a result of learning video generation in the pixel space. 

\noindent\textbf{Latent Diffusion for Video Generation.}
To reduce the complexity of video generation, latent-based models are explored~\cite{wu2022nuwa,hong2022cogvideo,villegas2023phenaki,wu2022tuneavideo,zhou2022magicvideo,he2022latent,esser2023structure}. Here we focus the discussions on latent diffusion models. Tune-A-Video~\cite{wu2022tuneavideo} finetunes a pretrained T2I model on a single video to enable one-shot video generation with the same action as the training video. Esser \textit{et al.}~\cite{esser2023structure} leverage monocular depth estimations and content representation to learn the reversion of the diffusion process in the latent space for video editing. The work that is most similar to ours is MagicVideo~\cite{zhou2022magicvideo}, where the authors use a T2I U-Net with a frame-wise adaptor and a directed temporal attention module for T2V generation. However, similar to~\cite{ho2022video,Ho2022ImagenVH,singer2023makeavideo}, these approaches need to use additional parameters to model the temporal dimension in videos. Our work can leverage the T2I U-Net as is for video generation, which is more efficient and enables both T2I and T2V generation in one unified framework.

\if, 0
Conventional approaches~\cite{zhu2019dm,tao2022df} focus on the model design on fixed datasets, such as MS COCO~\cite{lin2014microsoft}, CUB~\cite{welinder2010caltech}. 
While recent approaches achieve competitive performance compared to the previous domain-specific models in a zero-shot fashion.
In more detail, CogView~\cite{ding2021cogview}, N\"UWA~\cite{wu2022nuwa}, DALLE~\cite{ramesh2021zero} and Parti~\cite{yu2022scaling} train an auto-regressive transformer on hundred million to billion text-image pairs. GLIDE~\cite{nichol2021glide} and DALLE2~\cite{ramesh2022hierarchical} focus on large-scale training on diffusion-based models and achieve higher quality results. However, both works are trained in pixel space and require additionally trained super-resolution models to achieve higher resolution. One solution is proposed by latent diffusion~\cite{rombach2022high}, which directly generates high-resolution images by learning from latent space and thus reduces the computational cost.
Our work follows the idea of latent diffusion and extends it to text-to-video generation.

\noindent\textbf{Text-To-Video Generation.}
Similar to the evolution in text-to-image generation, conventional works in video generation also focus on the model design on fixed datasets, including GAN-based approaches~\cite{TGAN2020}, transformer-based approaches~\cite{yan2021videogpt,wu2021godiva}, diffusion-based approaches~\cite{ho2022video}, etc.
Comparing all generative models, diffusion models gain significant interest due to their high-quality generated results.
However, different from images, collecting large-scale, high-quality text-video pairs is difficult, expensive, and time-consuming.
A natural idea is to use the pretrained image model to facilitate spatial semantic learning in video generation. 
Therefore, some recent approaches finetune pretrained text-to-image models on text-video pairs, including transformer-based methods, such as CogVideo~\cite{hong2022cogvideo} and Phenaki~\cite{anonymous2023phenaki}, and diffusion-based methods, such as Make-A-Video~\cite{singer2023makeavideo} and Imagen Video~\cite{Ho2022ImagenVH}. 
In terms of existing diffusion-based methods~\cite{ho2022video,singer2023makeavideo,Ho2022ImagenVH}, a typical pipeline is first to generate a low-resolution video and then upsample it to a higher resolution with super-resolution models. models can generate high-quality videos in this way, however, are lack efficiency.
In this paper, we focus on improving the efficiency of generating high-quality videos.

\fi
\section{Method}
\label{sec:method}

This section introduces our~\latentshift that extends a latent diffusion model (LDM) from T2I generation to T2V generation through the temporal shift module. 
In Sec.~\ref{sec:stable}, we introduce the background of the LDM for T2I generation. 
Sec.~\ref{sec:temporal_shift} shows the mechanism, rationale, and effects of the temporal shift module.
Sec.~\ref{sec:shift} gives an overview of the proposed~\latentshift for T2V generation.

\subsection{Latent Image Diffusion Models}
\label{sec:stable}
There are two training stages in the latent image diffusion models: 
1) an autoencoder is trained to compress images into compact latent representations; 
2) a diffusion model based on the U-Net architecture is trained on text-image pairs to learn T2I generation in the latent space. 

\noindent\textbf{Latent Representation Learning.}
The latent space is learned by an autoencoder that consists of an encoder and a decoder. 
Given an RGB image $\mathbf{x}\in\mathbb{R}^{H\times W\times 3}$, the encoder $\mathcal{E}$ first compresses $\mathbf{x}$ into a latent representation $\mathbf{z}=\mathcal{E}(\mathbf{x})\in\mathbb{R}^{h\times w\times c}$ and then the decoder $\mathcal{D}$ reconstructs the image $\widetilde{\mathbf{x}}=\mathcal{D}(\mathbf{z})\in\mathbb{R}^{H\times W\times 3}$ from $\mathbf{z}$, where $H$ and $W$ represent the image height and width in the pixel space, $h$, $w$ and $c$ represent the height, width, and channel size of the feature maps in the latent space. 
The encoder downsamples the image by a factor of $f=H/h=W/w=2^m$ with $m\in\mathbb{N}$. 
VQGAN~\cite{esser2021taming} and VAE~\cite{kingma2014auto} are two widely used architectures of the autoencoder. We use a pretrained VAE model in our work.

\noindent\textbf{Conditional Latent Diffusion Models.}
Diffusion models~\cite{ho2020denoising,nichol2021improved} are generative models that are learned to recursively denoise from a normal distribution to a data distribution. There are different ways to parameterize the model. It can be trained by adding noise to the data and estimating the noise at different time steps. Specifically, given an image $\mathbf{x}$ that is encoded to the latent space $\mathbf{z}$, we add Gaussian noise into $\mathbf{z}$ defined as: 
\begin{equation}
\begin{aligned}
    \mathbf{z}_t=&\alpha_t\mathbf{z}_{0}+\sigma_t\epsilon, \epsilon\sim\mathcal{N}(0,1), \\
\end{aligned}
\label{eq:q_step}
\end{equation}
 where $\alpha_t$ and $\sigma_t$ are functions of $t$ following the definition in~\cite{ho2022video} that control the noise schedule, $t$ is the diffusion step that is uniformly sampled from $\{1,\dots,T\}$ during training where $T$ is the total number of time steps. $\mathbf{z}_0=\mathbf{z}$ is the original latent space before adding noise. 

The T2I LDM is trained on text-image pairs ($\mathbf{x}$, $\mathbf{y}$). The text $\mathbf{y}$ is encoded through a text encoder $\mathcal{C}$ to a representation $\mathcal{C}(\mathbf{y})$, which is mapped to the U-Net's spatial attention layers through the cross attention scheme.
The conditional latent diffusion model is trained to estimate the noise $\epsilon$ given a noisy input and conditioned on the text representation. A mean squared error loss is used:
\begin{equation}
    \mathcal{L}_{img} = \mathbb{E}_{\mathcal{E}(\mathbf{x}),\mathbf{y},\epsilon,t}\left[\left\|\epsilon-\epsilon_\theta(\mathbf{z}_{t},t,\mathcal{C}(\mathbf{y}))\right\|_2^2\right],
\end{equation}
\begin{figure}[t]
    \centering
    \vspace{-3mm}
    \includegraphics[width=\linewidth]{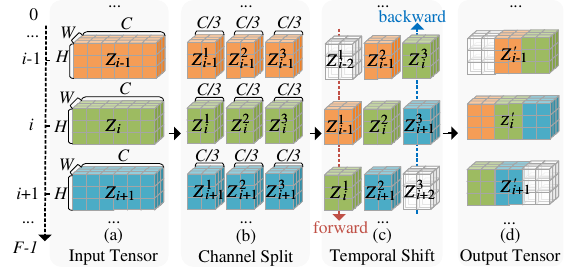}
    \vspace{-3mm}
    \caption{An illustration of temporal shift. $C$, $H$, $W$, $F$ represent channel, height, width, and frame, respectively. 
    }
    \label{fig:shift}
\end{figure}
\begin{figure*}
    \centering
    \includegraphics[width=\textwidth]{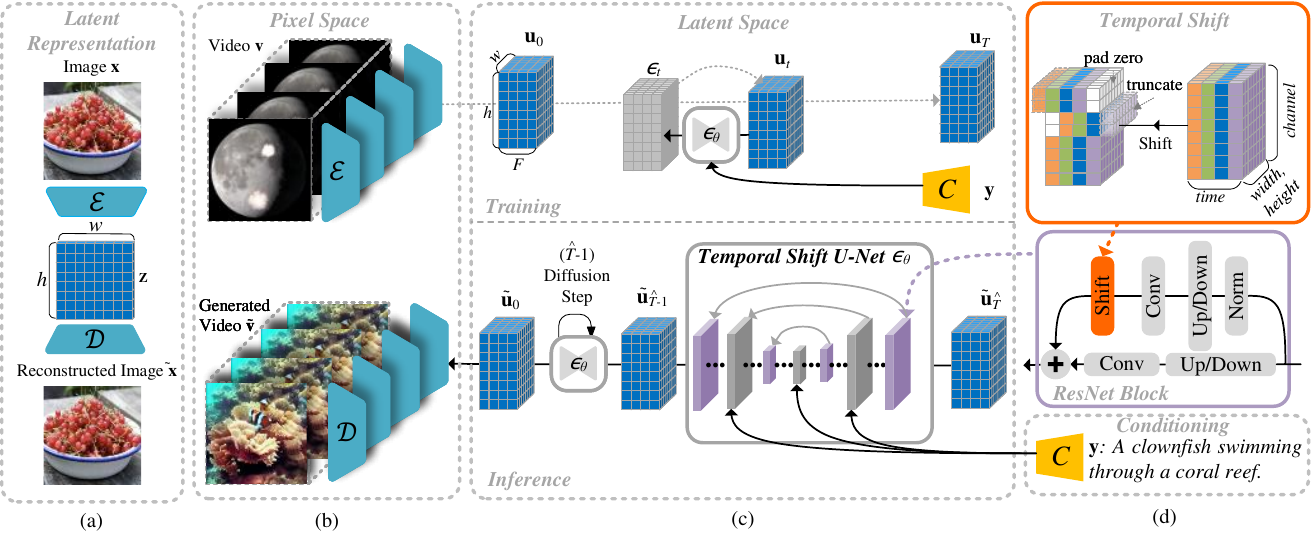}
    \caption{An illustration of our framework. From left to right: \textbf{(a)} An autoencoder is trained on images to learn latent representation. \textbf{(b)} The pretrained autoencoder is adapted to encode and decode video frames independently. \textbf{(c)} During training, a temporal shift U-Net $\epsilon_\theta$ learns to denoise latent video representation at a uniformly sampled diffusion step $t\in [1,T]$. During inference, the U-Net gradually denoises from a normal distribution from step $\hat{T}-1$ to $0$ where $\hat{T}$ is the number of resampled diffusion steps in inference. \textbf{(d)} The U-Net $\epsilon_\theta$ is composed of two key building blocks: the $2$D ResNet blocks with convolutional layers, highlighted in \textcolor{violet}{\bf\textit{violet}}, and the transformer blocks with spatial attention layers, colored in \textcolor{gray}{\bf{\textit{gray}}}. The temporal shift module, highlighted in \textcolor{red}{\bf{\textit{red}}}, shifts the feature maps along the temporal dimension. It is inserted into the residual branch of each $2$D ResNet block. The text condition is applied to the transformer blocks via cross-attention. The channel dimension $c$ in the latent space representation of $\mathbf{z}$ and $\mathbf{u}$ are omitted for clarity.
    }
    \label{fig:pipeline}
\end{figure*}

\subsection{Temporal Shift} 
\label{sec:temporal_shift}
Motivated by~\cite{lin2019tsm,munoz2021temporal}, the temporal shift operation can leverage a $2$D spatial network to handle both spatial and temporal information by mingling the information from neighboring frames with the current frame. 
Figure~\ref{fig:shift}
illustrates the temporal shift operation with an example of three frames. Let $Z\in \mathbb{R}^{C \times F\times H\times W}$ be the input to a temporal shift module, where $Z_i\in \mathbb{R}^{C\times H \times W}$ is the feature map for the $i^{\mathrm{th}}$ frame. We first split each $Z_i$ into $Z_i^1$, $Z_i^2$, and $Z_i^3$ along the channel dimension $C$, where $Z_i^j\in \mathbb{R}^{\frac{C}{3}\times H \times W}$. Then we shift $Z_i^1$ forward and $Z_i^3$ backward along the temporal (frame) dimension $F$. Finally, we merge the temporal-shifted features together. The output of the temporal shift module for the $i^{\mathrm{th}}$ frame is:
\begin{equation}
\small
Z'_i =
\begin{cases}
    [\mathbf{0}, Z_0^2, Z_1^3] & i=0, \\
    [Z_{i-1}^1, Z_i^2, Z_{i+1}^3] & 0<i<F-1,\\
    [Z_{F-2}^1, Z_{F-1}^2, \mathbf{0}] & i=F-1.\\
\end{cases}
\end{equation}
Here $\mathbf{0}$ denotes zero-padded feature maps.

The temporal shift module enables each frame's feature $Z_i$ to contain the channels of the adjacent frames $Z_{i-1}$ and $Z_{i+1}$ and thus enlarge the temporal receptive field by $2$.
The $2$D convolutions after the temporal shift, which operate independently on each frame, 
can capture and model both the spatial and temporal information as if running an additional $1$D convolution with a kernel size of $3$ along the temporal dimension~\cite{lin2019tsm}.

\subsection{Latent-Shift for T2V Generation} 
\label{sec:shift}

We adopt a pretrained autoencoder and a U-Net latent diffusion model. 
The autoencoder is fixed to encode and decode videos independently for each frame. We finetune the U-Net with the added temporal shift modules to enable video modeling for T2V generation.

The pretrained U-Net comprises two key building blocks: 1) $2$D ResNet blocks that consist of mainly convolutional layers, and 2) spatial transformer blocks that mainly include attention layers; both are designed only to model the spatial relationships. It is essential to enable the U-Net to model temporal information between video frames to learn meaningful motion. One straightforward direction is to add additional layers, as widely used in prior work. 
For example, VDM~\cite{ho2022video} and Magic Video~\cite{zhou2022magicvideo} add a temporal attention layer after each spatial attention layer. Make-A-Video~\cite{singer2023makeavideo} adds $1$D convolutional layers in the ResNet blocks and temporal attention layers in the transformer blocks. While it is intuitive to add new layers to extend the U-Net from modeling images to videos, 
we explore ways to use the U-Net as is for video generation.

To this end, we propose to incorporate the aforementioned temporal shift modules into the U-Net for T2V generation.
Our framework is illustrated in Fig.~\ref{fig:pipeline}. Specifically, we insert a temporal shift module inside the residual branch, which shifts the feature maps along the temporal dimension with zero padding and truncation, as shown in Fig.~\ref{fig:pipeline} (d).

Given a video $\mathbf{v}\in\mathbb{R}^{F\times H\times W\times 3}$, where $F$ represents the number of frames, we use the pretrained encoder $\mathcal{E}$ to encode each frame to get the latent video representation as $\mathbf{u}\in\mathbb{R}^{F\times h\times w\times c}$. The diffusion model is learned on this encoded latent space. We use a pretrained text encoder and add the text representation the same way in the attention layers as a condition. Similar to prior work, we also use classifier-free guidance~\cite{ho2021classifier,nichol2021glide} to improve sample fidelity to the input text. This is enabled by randomly dropping out the text input with a certain probability to learn unconditional video denoising during training.
Similar to the diffusion process for image generation, we add noise $\epsilon$ into $\mathbf{u}$ at each training time step as:
\begin{equation}
\mathbf{u}_t=\alpha_t\mathbf{u}_{0}+\sigma_t\epsilon, \epsilon\sim\mathcal{N}(0,1),
\end{equation}
where $\alpha_t$, $\sigma_t$, $t$ are defined the same way as in Eqn.~\ref{eq:q_step}. $\mathbf{u}_0=\mathbf{u}$ is the initial latent video representation before adding noise. The training objective is to estimate the added noise from the noisy input, which is defined as:
 \begin{equation}
    \mathcal{L}_{vid} = \mathbb{E}_{\mathcal{E}(\mathbf{x}),\mathbf{y},\epsilon,t}\left[\left\|\epsilon-\epsilon_\theta(\mathbf{u}_{t},t,\mathcal{C}(\mathbf{y}))\right\|_2^2\right],
\end{equation}
where $\theta$ denotes the learnable parameters from the pre-trained T2I U-Net model and $\mathcal{C}$ is the pretrained text encoder which is fixed during training.

Diffusion models are typically trained on a large number of discrete time steps ({\it{e.g.}}, $1000$) but can be used to sample data with fewer time steps to improve efficiency during inference. 
We use the DDPM sampler~\cite{ho2020denoising} with classifier-free guidance~\cite{ho2021classifier} and conduct sampling with $\hat{T}=100$ steps.

\begin{figure*}[t]
    \centering
    \includegraphics[width=\textwidth]{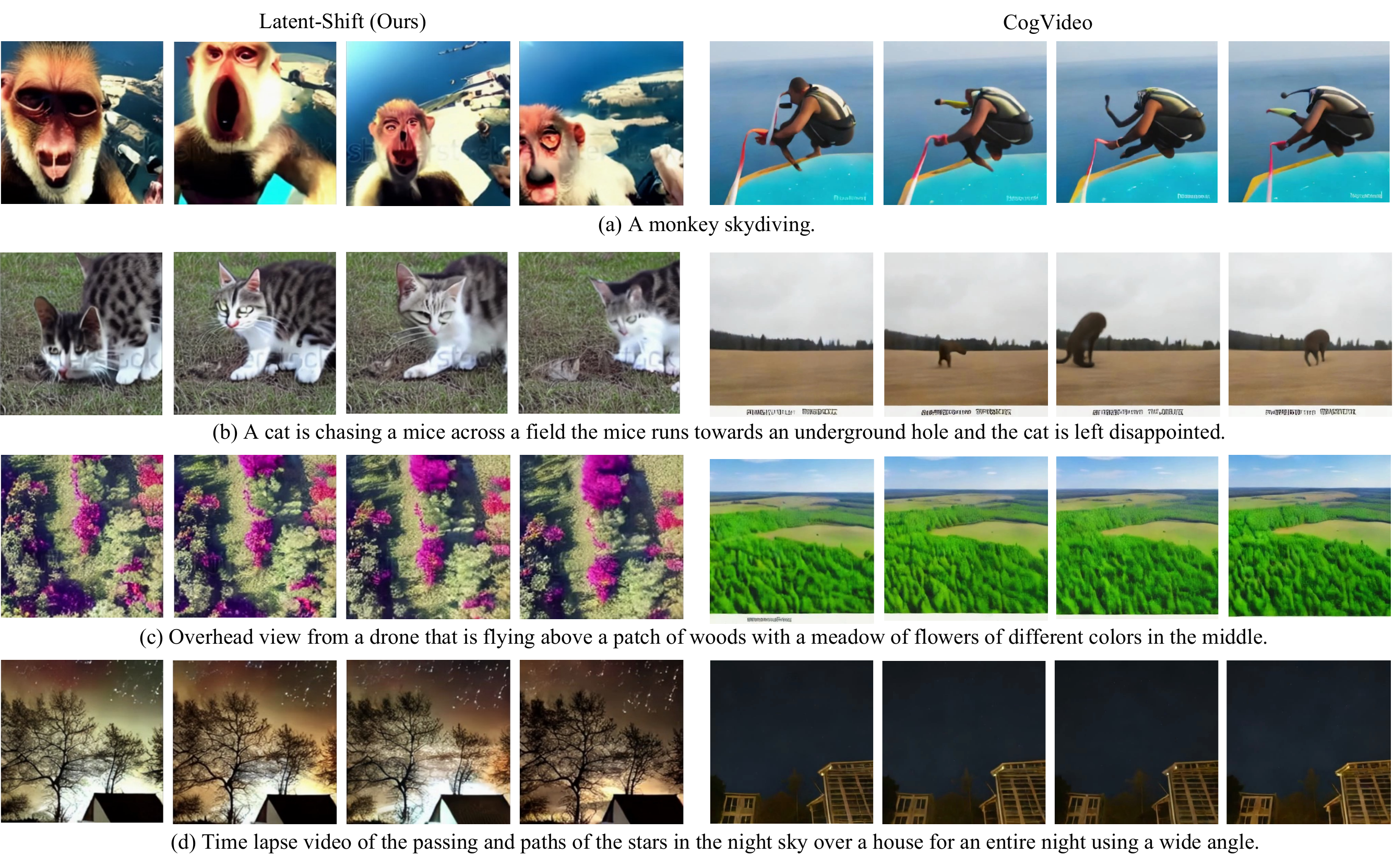}
    \vspace{-4mm}
    \caption{Text-to-Video generation comparison with CogVideo~\cite{hong2022cogvideo} on the user study evaluation set. Our model can generate more semantically correct content with meaningful motions.}
    \label{fig:free_form_comp}
\end{figure*}

\begin{figure*}[t]
    \centering
    \includegraphics[width=0.96\textwidth]{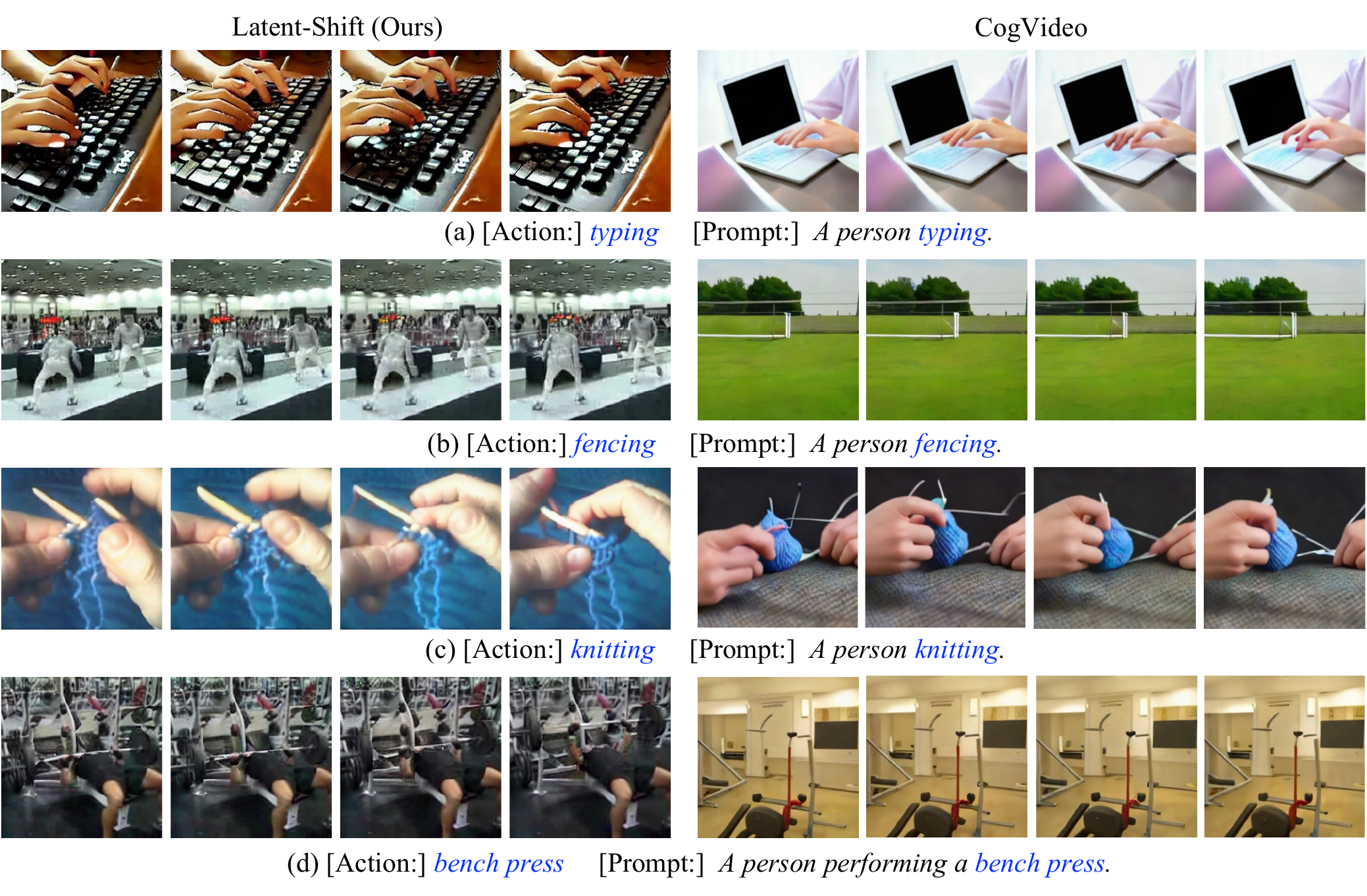}
    \vspace{-1mm}
    \caption{Text-to-video generation comparison with CogVideo~\cite{hong2022cogvideo} on the UCF101~\cite{soomro2012ucf101} dataset.}
    \label{fig:ucf_comp}
\end{figure*}

\section{Experiments}
\label{sec:experiments}

\subsection{Implementation Details}
Training is conducted on the WebVid~\cite{bain2021frozen} dataset with $10$M text-video pairs. Following prior work, we report results on UCF-101~\cite{soomro2012ucf101}, MSR-VTT~\cite{xu2016msr} with commonly used metrics including Inception Score (IS), Fréchet Image Distance (FID), Fréchet Video Distance (FVD), and CLIP similarity (CLIPSIM) between the generated video frames and the text. In addition, we conduct a user study comparing to CogVideo~\cite{hong2022cogvideo} 
via video quality and text-video faithfulness metrics. For all evaluations, we generate a random sample for each text without any automatic ranking. More details on hyperparameter settings are available in the supplementary materials.

\begin{table}[]
    \centering
    \caption{Zero-Shot T2V generation comparison on MSR-VTT. The results of CogVideo are cited from~\cite{singer2023makeavideo}. The best and second-best results are marked in bold and underlined, respectively.}
    \begin{tabular}{@{}lc|cc@{}}
    \toprule
    Method & Zero-Shot & FID $\downarrow$ & CLIPSIM $\uparrow$ \\
    \midrule
    GODIVA~\cite{wu2021godiva} & No & - & $0.2402$ \\
    N\"UWA~\cite{wu2022nuwa} & No & $47.68$ & $0.2439$\\
    CogVideo~\cite{hong2022cogvideo} & Yes & $23.59$ & $0.2631$ \\
    Make-A-Video~\cite{singer2023makeavideo} & Yes & $\bf 13.17$ & $\bf 0.3049$\\
    \midrule
    Latent-VDM & Yes & $\underline{14.25}$ & $0.2756$\\
    \latentshift (ours) & Yes & $15.23$ & $\underline{0.2773}$\\
    \bottomrule
    \end{tabular}
    \label{tab:MSRVTT}
\end{table}

\subsection{Main Results}
\noindent\textbf{Evaluation on MSR-VTT.} We conduct a zero-shot evaluation on the MSR-VTT test set. Following prior works~\cite{wu2022nuwa}, we use all the captions in the test set and calculate frame-level metrics. We compare \latentshift with prior works that are evaluated on MSR-VTT. In addition, we also implement a baseline of latent video diffusion model with the widely used temporal attention, termed as Latent-VDM.
Both \latentshift and Latent-VDM are trained with the same setting. The results are shown in Tab.~\ref{tab:MSRVTT}.
The performance of \latentshift is competitive with prior works. In most cases, it already outperforms several methods with noticeable margins. Even though \latentshift does not outperform Make-A-Video due to our limited model size (see Tab.~\ref{tab:model_components}), the performance is much closer than other models.  

\begin{table}[]
    \centering
    \vspace{-1mm}
    \caption{T2V generation comparison on UCF-101.}
    \begin{tabular}{l|cc}
    \toprule
     Method  & IS $\uparrow$ & FVD $\downarrow$ \\
    \midrule
    VideoGPT\cite{yan2021videogpt} & $24.69\pm0.3$ & - \\
     TGANv2~\cite{saito2020train} & $26.60 \pm 0.47$ & - \\
     DIGAN~\cite{yu2022digan} & $32.70 \pm 0.35$ & $577\pm 22$ \\
     DVD-GAN\cite{clark2019adversarial} & $32.97\pm1.7$ & - \\
     MoCoGAN-HD~\cite{tian2021a} & $33.95 \pm 0.25$ & $700\pm 24$ \\
     CogVideo~\cite{hong2022cogvideo}  & $50.46$ & 626 \\
     VDM~\cite{ho2022video} & $57.80 \pm 1.3$ & - \\
     TATS-base~\cite{ge2022long}  & $79.28 \pm 0.38$ & $\underline{278\pm 11}$ \\
     Make-A-Video~\cite{singer2023makeavideo}  & $82.55$ & $\bf 81.25$ \\
     \midrule
     Latent-VDM & $\underline{90.74}$ & $358.34$ \\
     \latentshift (ours) & $\bf 92.72$ & $360.04$ \\
     \bottomrule
    \end{tabular}
    \label{tab:UCF101}
\end{table}

\begin{table}[]
    \centering
    \vspace{-1mm}
    \caption{User study results. The numbers show the percentages of raters who prefer our \latentshift or Latent-VDM over CogVideo~\cite{hong2022cogvideo}.}
    \begin{tabular}{l|cc}
    \toprule
    Method & Quality (\%) & Faithfulness (\%)\\
    \midrule
    Latent-VDM  & $56.4$ & $53.8$ \\
   \latentshift (ours) & $\bf 58.0$ & $\bf 62.9$\\
    \bottomrule
    \end{tabular}
    \label{tab:human_eval}
\end{table}

\begin{table*}[]
\centering
    \caption{Model size and inference speed comparisons. The speed is measured in seconds on one A100 (80GB) GPU. (For CogVideo, there are $6$B parameters that are shared among the T2V model and the frame interpolation model.)}
    \resizebox{\linewidth}{!}{
    \setlength{\tabcolsep}{1mm}
    \begin{tabular}{l|c|c|c|c|c|c|c|c}
    \toprule
    \multirow{2}*{Method} & \multicolumn{7}{c|}{Parameters (Billion)} & \multirow{2}*{Speed (s)} \\ \cline{2-8}
    & T2V Core & Auto Encoder & Text Encoder & Prior Model & Super Resolution & Frame Interpolation & Overall & \\
    \midrule
    CogVideo~\cite{hong2022cogvideo} & $7.7$ & $0.10$ & $-$ & $-$ & $-$ & $7.7$ & $15.5$ & $434.53$ \\ 
    Make-A-Video~\cite{singer2023makeavideo} & $3.1$ & $-$ & $0.12$ & $1.3$ & $1.4+0.7$ & $3.1$ & $9.72$ & $-$ \\
    Imagen Video~\cite{Ho2022ImagenVH} & $5.6$ & $-$ & $4.6$ & $-$ & $1.2+1.4+0.34$ & $1.7+0.78+0.63$ & $16.25$ & $-$ \\ \hline
    Latent-VDM & $0.92$ & $0.08$ & $0.58$ & $-$ & $-$ & $-$ & $\underline{1.58}$ & $\underline{28.62}$ \\
    \latentshift (ours) & $0.87$ & $0.08$ & $0.58$ & $-$ & $-$ & $-$ & $\bf1.53$ & $\bf 23.40$ \\
    \bottomrule
    \end{tabular}
    }
    \vspace{-1mm}
    \label{tab:model_components}
\end{table*}

\noindent\textbf{Evaluation on UCF-101.} We evaluate the performance on UCF-101 by finetuning on the dataset. The UCF-101 dataset consists of $13,320$ videos\footnote{Following prior work, we train on all the samples from both train and test splits, and evaluate on the train split.} from $101$ human action labels. We construct templated sentences for each class to form a text prompt. Then we finetune our pretrained T2V model to fit the UCF-101 data distribution. During inference, we perform class-conditional sampling to generate videos with the same class distribution as the training set for evaluation, following~\cite{singer2023makeavideo}. As shown in Tab.~\ref{tab:UCF101}, our approach achieves state-of-the-art results on IS and a competitive score on FVD. 

\noindent\textbf{User Study.} 
It is well known that automatic evaluation metrics are far from perfect. 
Therefore, it is more desirable to conduct user studies. To this end, we use the evaluation set from~\cite{singer2023makeavideo} that consists of $300$ text prompt collected from Amazon Mechanical Turk (AMT). 
We compare to CogVideo~\cite{hong2022cogvideo} and evaluate both video quality and text-video faithfulness. The user study is conducted on AMT where $5$ different raters evaluate each comparison and the majority vote is taken. 

The results are shown in Tab.~\ref{tab:human_eval}. Our approach achieves better results in both video quality and text-video faithfulness compared to CogVideo. This is consistent with the automatic evaluations. As shown in Tab.~\ref{tab:model_components}, our model is also much more efficient than CogVideo.

\begin{table}[]
    \centering
    \caption{Evaluation on MSR-VTT in the zero-shot setting. We use LDM and our model to generate images and treated them as frozen videos for this comparison.}
    \begin{tabular}{l|cc}
    \toprule
    Method & FID $\downarrow$ & CLIPSIM $\uparrow$ \\
    \midrule
    LDM & $\underline{15.36}$ & $\bf 0.2910$ \\
   \latentshift (T2I) & $15.64$ & $0.2737$ \\
   \latentshift & $\bf 15.23$ & $\underline{0.2773}$\\
    \bottomrule
    \end{tabular}
    \vspace{-2mm}
    \label{tab:image-video}
\end{table}

\noindent\textbf{Model Size and Inference Speed.}
We compare the model size and inference speed in Tab.~\ref{tab:model_components}. Only Cogvideo is chosen for speed comparison since it is the only open-sourced zero-shot T2V model.
Latent-Shift is much smaller than prior works and much faster than CogVideo. Without a large number of parameters,~\latentshift achieves better results than CogVideo in various benchmarks. This validates the effectiveness of~\latentshift.

\subsection{Ablation Study}
\noindent\textbf{Temporal Shift {\it v.s.} Temporal Attention.} We compare our temporal shift module (\latentshift) with the widely used temporal attention layers (Latent-VDM) in the U-Net extension from image to video modeling. As already shown in Tabs.~\ref{tab:MSRVTT}, ~\ref{tab:UCF101}, ~\ref{tab:human_eval},
\latentshift performs better than Latent-VDM in most cases, especially with a large margin in the user study. 
Furthermore, \latentshift requires fewer model parameters and thus enables relatively faster inference than Latent-VDM, as shown in Tab.~\ref{tab:model_components}.

\noindent\textbf{Image Generation as a Frozen Video.} Our finetuned \latentshift can be used for both image and video generation where an image can be considered as a frozen video with a single frame. In Tab.~\ref{tab:image-video}, we compare \latentshift with LDM on MSR-VTT. We observe that after training \latentshift for T2V generation, it can still perform reasonable T2I generation. However, this also suggests that the metrics on MSR-VTT evaluation is not ideal as they do not account for the motion information in the videos. A better metric for the automatic evaluation of zero-shot T2V generation is needed.

\begin{figure}[t]
    \centering
    \includegraphics[width=0.49\textwidth]{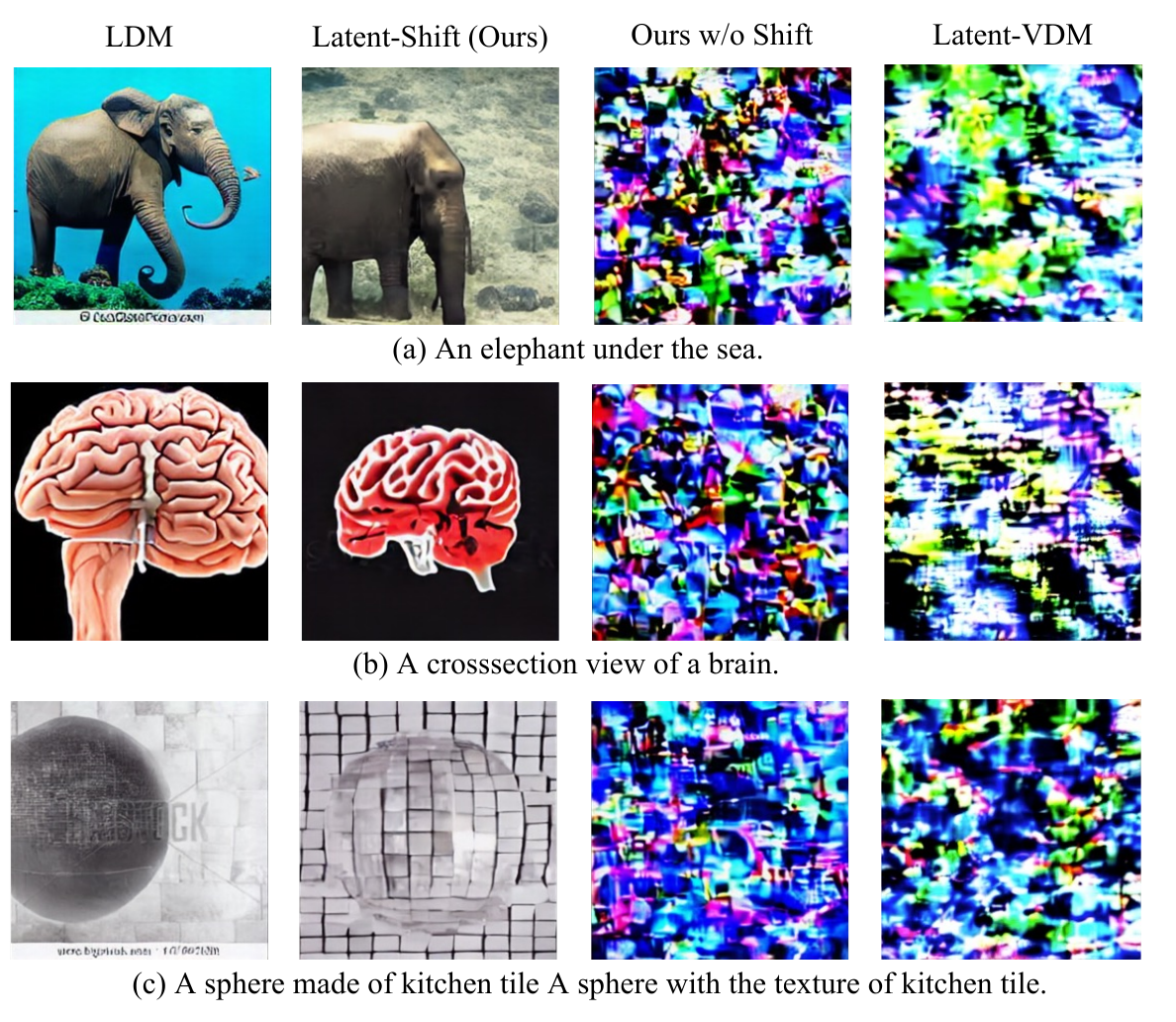}
    \vspace{-4mm}
    \caption{T2I generation comparison. \latentshift can generate meaningful images although it is finetuned for video generation but the latent-VDM with temporal attention cannot. Our method fails to generate images if the temporal shift module is removed. This demonstrates the shift module's importance even though adding temporal shift on a single image means dropping a part of the feature maps.}
    \label{fig:t2i}
\end{figure}

\begin{figure}[t]
    \centering
    \includegraphics[width=0.47\textwidth]{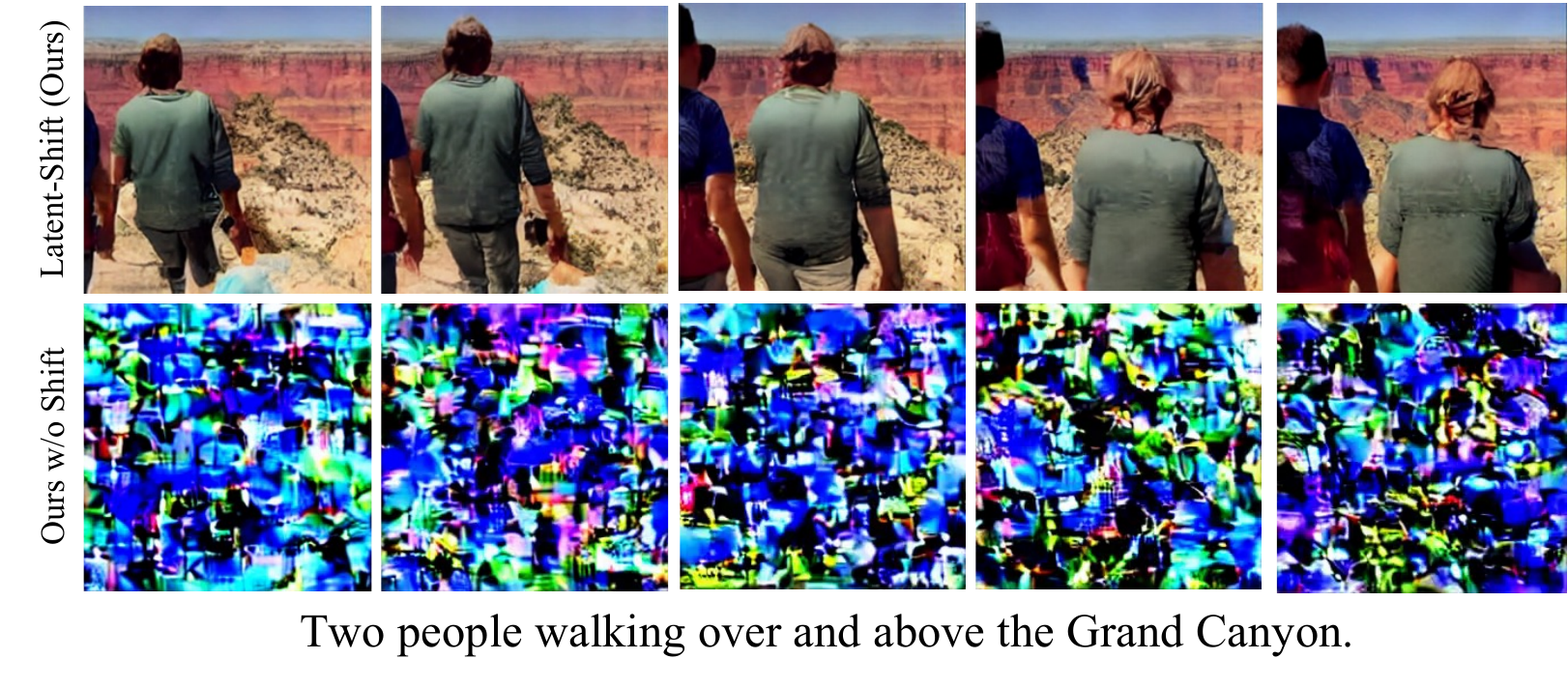}
    \vspace{-1mm}
    \caption{T2V generation with and without temporal shift during inference. Our model would not generate meaningful videos if the temporal shift module is disabled.}
    \label{fig:t2v-no-shift}
\end{figure}

\begin{figure}[t]
    \centering
    \includegraphics[width=0.49\textwidth]{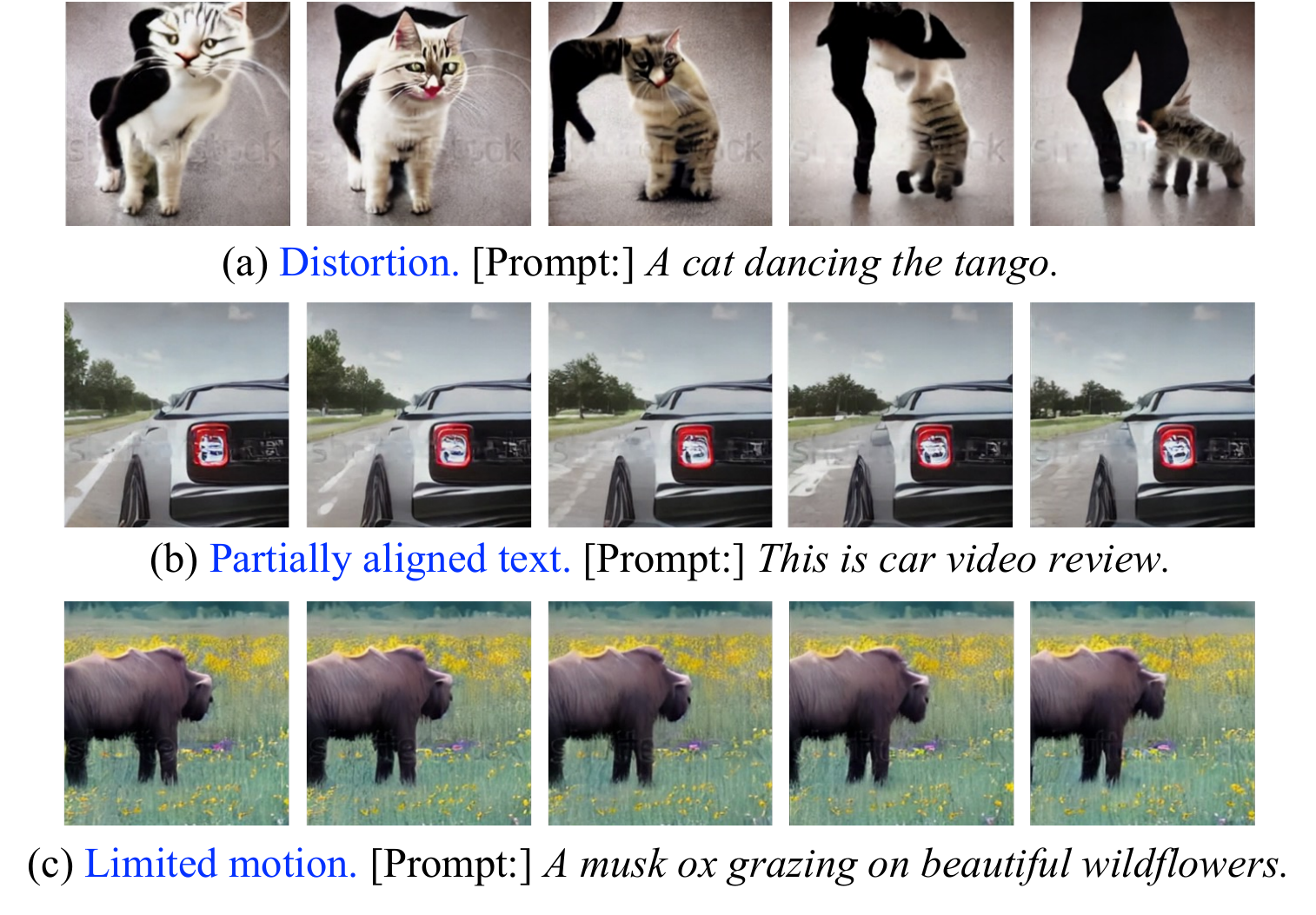}
    \vspace{-4mm}
    \caption{Failure cases of our approach. (a) Content distortion when generating videos with complex objects and action compositions. (b) Partially aligned text where not everything mentioned in the text can be generated. (c) Limited motion when the action is very subtle in the input text.}
    \label{fig:failure}
\end{figure}

\subsection{Qualitative Results}
\noindent\textbf{T2V Generation.} We show visual comparisons with CogVideo in Figs.~\ref{fig:free_form_comp} and ~\ref{fig:ucf_comp} for different evaluation sets. 

In both cases, \latentshift can generate semantically richer content with a meaningful motion that is faithful to the input text. This validates the effectiveness of our approach. 

\noindent\textbf{T2I and T2V Generation without Temporal Shift.} The temporal shift module is parameter-free, {\textit{i.e.,}} the U-Net for T2V generation is with the same parameters as the T2I model that it is initialized from. In more detail, for Latent-VDM, all context information from all frames is always available for each individual frame during training. Therefore, the model collapses during T2I inference with a single frame due to the lack of context information from the missing frames and the relative position inputs (column $4$). We also tried to remove the temporal attention layer during inference, but it does not help to enable T2I generation.
In contrast, in training \latentshift, the succeeding convolutional layers of the temporal shift module learn local context only from the previous and the next frames (all convolutional kernels are set to $3$). Meanwhile, the padded zeros in the first frame and the last frame enable the kernels to learn generation with missing context during training. Therefore, when we use \latentshift for T2I generation, it can still generate reasonable images (column $2$). Noted that the temporal shift module is necessary for T2I generation and cannot be removed, since the padded zeros indicate whether the context information is missing or not (Column $2$ \textit{vs} Column $3$). 

Similarly, we perform T2V generation with and without the temporal shift module. As shown in Fig.~\ref{fig:t2v-no-shift}, the video generation fails if not adding the temporal shift module.

\noindent\textbf{Failure Cases.} \latentshift works well for most text inputs but can struggle with some. We have observed three main types of failure cases (which are common), as shown in Fig.~\ref{fig:failure}. 
First, there might be artifacts of object distortion and frame flickering. It happens when the text contains mixed concepts that are not commonly seen in the real world ((a)). It could be limited by the scale of the text-video training data and the fact that the VAE model is trained on images only. To learn the latent space from both image and video patches as~\cite{villegas2023phenaki} has the potential to alleviate this issue. 
Second, ~\latentshift may not always generate videos that match the text exactly. There may be missing contents ((b)). 
Third, some generated videos will have limited motion. This is a common issue of T2V generation methods~\cite{ho2022video,singer2023makeavideo,zhou2022magicvideo}, which often happens when the action is subtle in the text ((c)).

\section{Conclusion}
\label{sec:conclusion}



In this paper, we present~\latentshift, a simple and efficient framework for T2V generation. We finetune a pre-trained T2I model with temporal shift on video-text pairs. The temporal shift module can model temporal information without adding any new parameters. Our model also preserves the T2I generation capability even though it is finetuned on videos, which is a unique property compared to many existing methods. The experimental results on MSRVTT and UCF101, along with user studies, demonstrate the effectiveness and efficiency of our approach.

{\small\balance
\bibliographystyle{ieee_fullname}
\bibliography{ms}
}

\newpage
\appendix
\onecolumn
\begin{center}
  \textbf{
    \Large Latent-Shift: Latent Diffusion with Temporal Shift \\for Efficient Text-to-Video Generation \\\vspace{0.4em}-- Supplementary Material --} \\
\hspace{1cm}
\end{center}

\section{Hyperparameter Settings}
\label{sec:supp_config}

\begin{table*}[b!]
    \centering
    \caption{The hyper-parameter setting of our models. AE denotes the auto-encoder to encode and decode videos. Common, \latentshift and Latent-VDM indicate whether the hyper-parameter belongs to both models, \latentshift or Latent-VDM.}
    \setlength{\tabcolsep}{2mm}
    \begin{tabular}{lc|lc}
        \toprule
        Hyper-parameter (common) & Value & Hyper-parameter (common) & Value  \\
        \midrule
        Image Size & 256 & Num Frame & 16 \\
        Guidance Scale & 7.5 & Text Seq Length & 77 \\
        Text Encoder & BERTEmbedder & First Stage Model & AutoencoderKL \\
        AE Double $z$ & True & AE $z$ channel & 4 \\
        AE Resolution & 256 & AE In Channel & 3 \\
        AE Out Channel & 3 & AE Channel & 128 \\
        AE Channel Multiplier & $[1,2,4,4]$ & AE Num ResBlock & 2 \\
        AE Atten Resolution & $[]$ & AE Dropout & 0.0 \\
        Store EMA & True & EMA FP32 & True \\
        EMA Decay & 0.9999 & Diffusion In Channel & 4 \\
        Diffusion Out Channel & 4 & Diffusion Channel & 320 \\
        Conditioning Key & crossattn & Noise Schedule & quad \\
        Encoder Channel & 1280 & Atten Resolution & $[4,2,1]$ \\
        Num ResBlock & 2 & Channel Multiplier & $[1,2,4,4]$ \\
        Transformer Depth & 1 & Batch Size & 4 \\
        Learn Sigma & False & Diffusion Step & 1000 \\
        Timestep Respacing & 100 & Sampling FP16 & False \\
        Learning Rate & $1e^{-5}$ & Sample Scheduler & DDPM \\
        Num Head & 8 & & \\
        \bottomrule
        \toprule
        Hyper-parameter (\latentshift) & Value & Hyper-parameter (Latent-VDM) & Value  \\
        \midrule
        Attention Block Type & SpatialTransformer & Attention Block Type & SpatialTemporalTransformer \\
        Shift Fold & 3 & & \\
        \bottomrule
    \end{tabular}
    \label{tab:supp_config}
\end{table*}

For video data, we evenly sample $16$ frames from a two seconds clip. 
We perform image resizing and center cropping to $256\times256$. 
The latent space is $32\times32\times4$. 
To apply temporal shift on the feature maps of each frame, we keep $1/3$ channels from the previous frame, $1/3$ from the current frame, and $1/3$ from the subsequent frame. 
Adam~\cite{kingma2015adam} is used for optimization, the learning rate is set to $1\times10^{-5}$, the batch size is set to $256$, the number of diffusion steps $T$ is set to $1000$, and bounds $\beta_1$ and $\beta_T$ are set to $8.5\times 10^{-4}$ and $1.2\times 10^{-2}$. 
During inference, the number of sampling steps $\hat{T}$ is set to $100$, and the guidance scale $s$ is set to $7.5$.
Table~\ref{tab:supp_config} shows the hyper-parameter settings of our models.

\section{Text-to-Video Generation}
\label{sec:supp_t2v}

In this section, we compare our proposed~\latentshift with CogVideo~\cite{hong2022cogvideo} and VDM~\cite{ho2022video} qualitatively, as shown in Figure~\ref{fig:supp_t2v}. 
We use the text prompts collected from VDM's website~\footnote{\url{https://video-diffusion.github.io/}}.
Comparing all three methods, our generated videos contain richer content and thus with higher visual quality. 

\begin{figure*}[h!]
\centering
\begin{subfigure}{\linewidth}
\includegraphics[width=\linewidth]{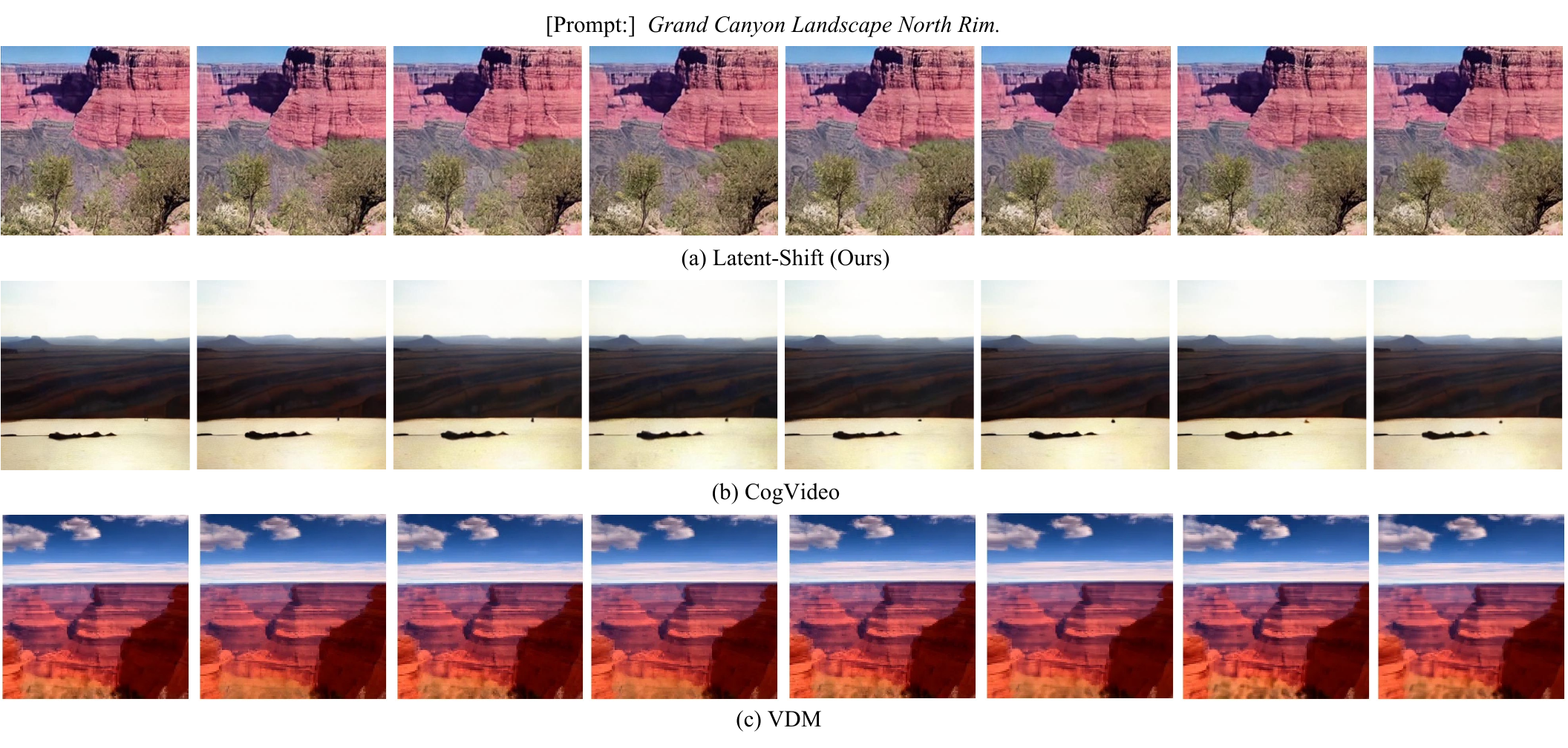}%
\end{subfigure}\hfill
\begin{subfigure}{\linewidth}
\includegraphics[width=\linewidth]{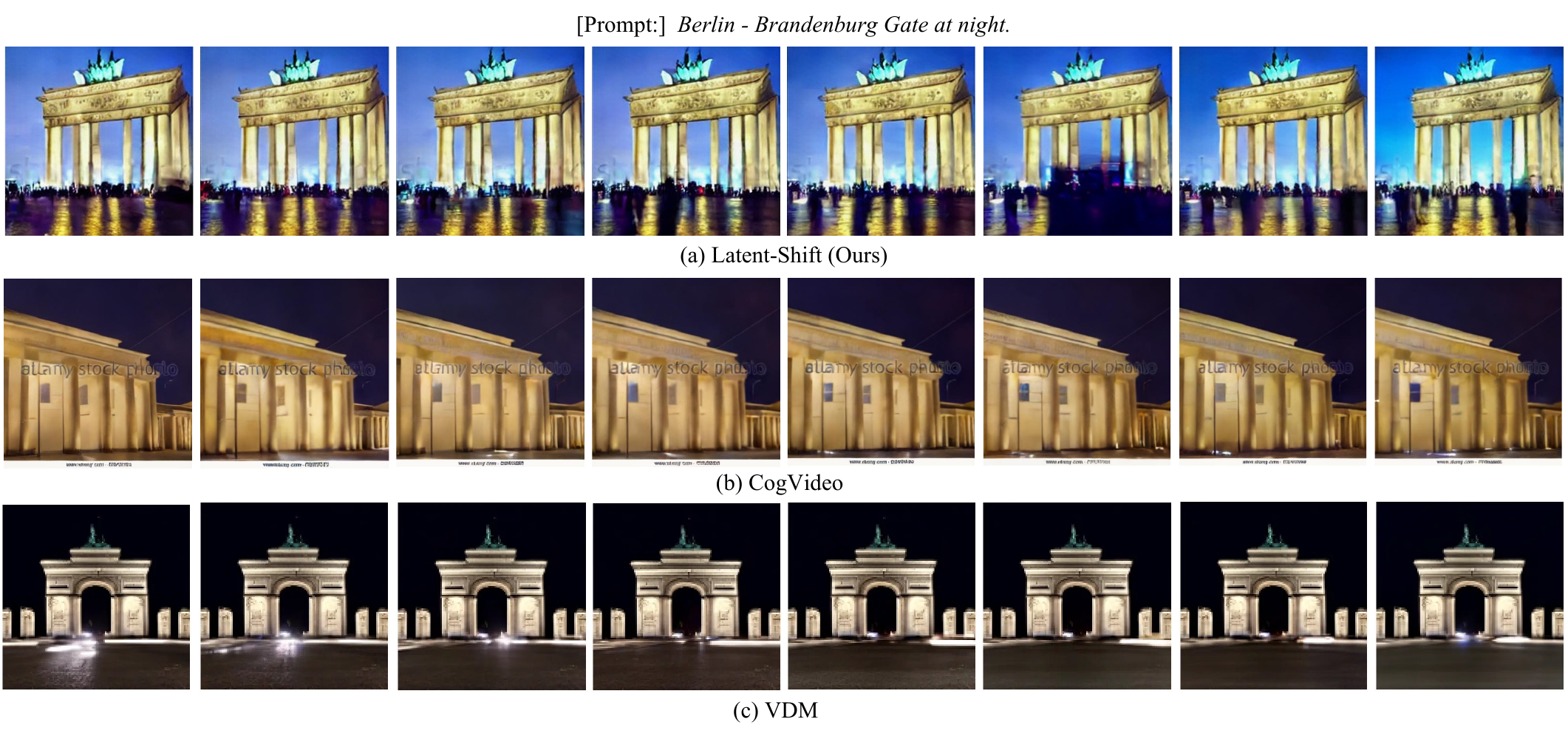}%
\end{subfigure}%
\caption{Text-to-Video generation comparison with CogVideo and Video Diffusion Models (VDM).}
\label{fig:supp_t2v}
\end{figure*}

\begin{figure*}
\continuedfloat
\centering
\begin{subfigure}{\linewidth}
\includegraphics[width=\linewidth]{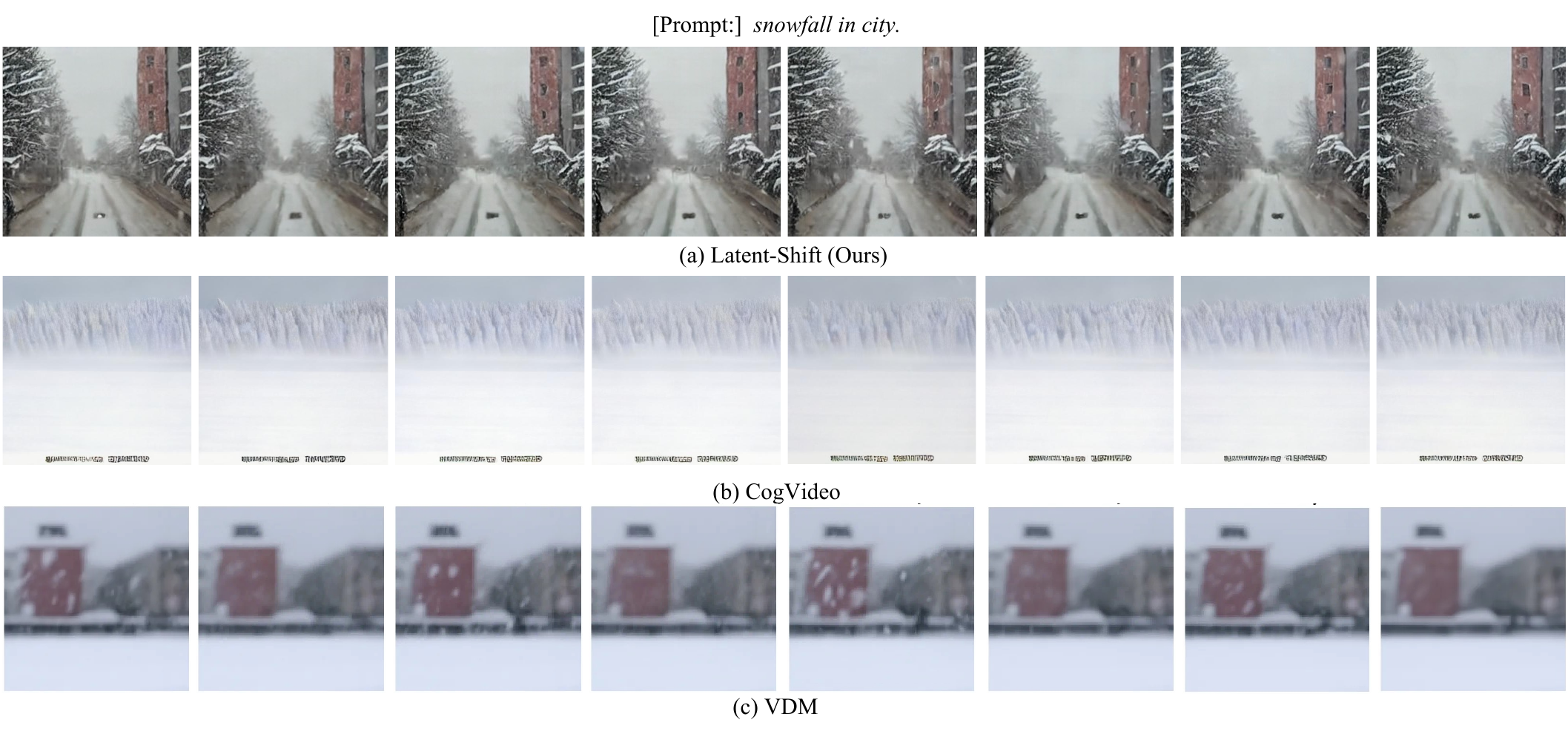}%
\end{subfigure}\hfill
\begin{subfigure}{\linewidth}
\includegraphics[width=\linewidth]{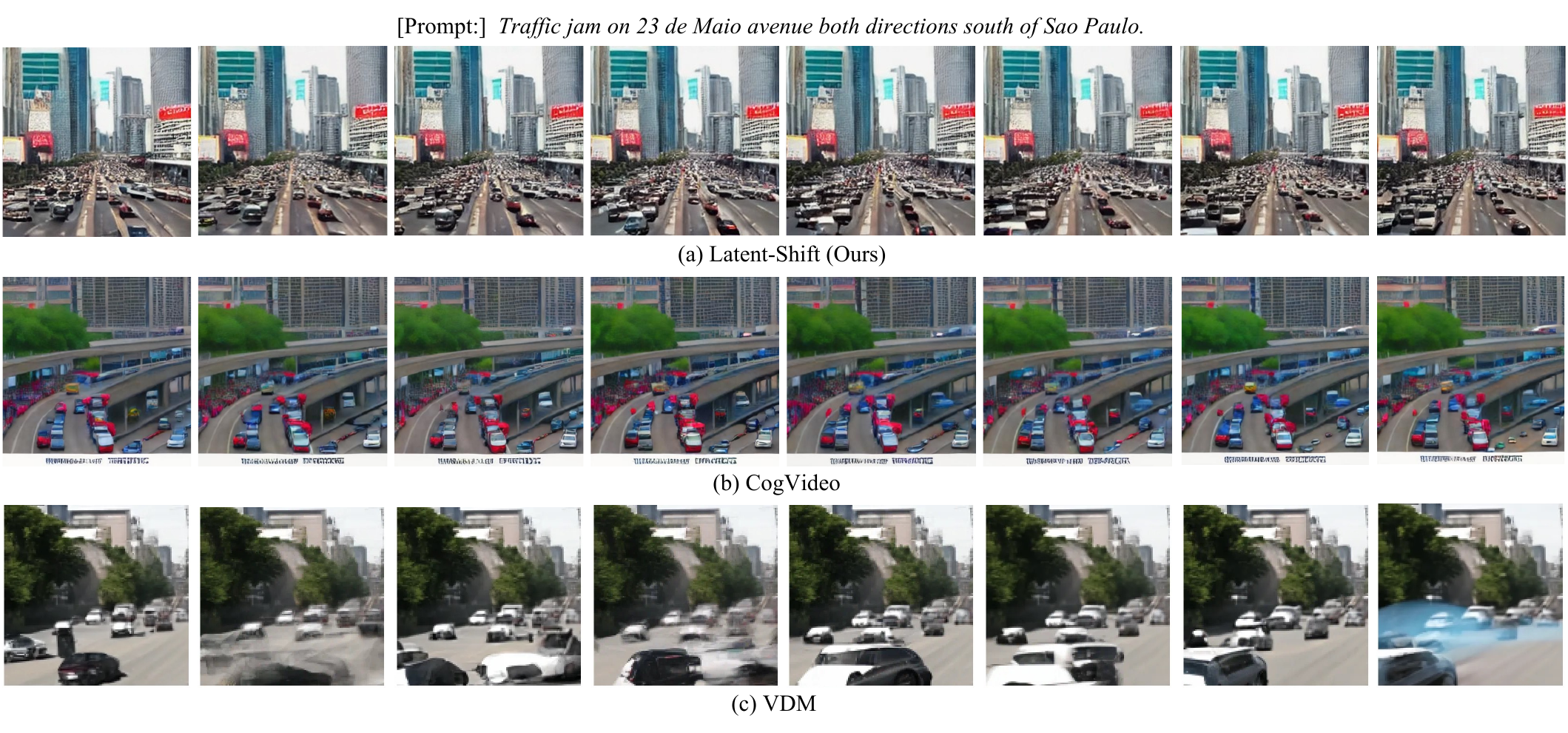}%
\end{subfigure}%
\caption{Text-to-Video generation comparison with CogVideo and Video Diffusion Models (VDM) - continued.}
\label{fig:supp_t2v}
\end{figure*}

\begin{figure*}
\continuedfloat
\centering
\begin{subfigure}{\linewidth}
\includegraphics[width=\linewidth]{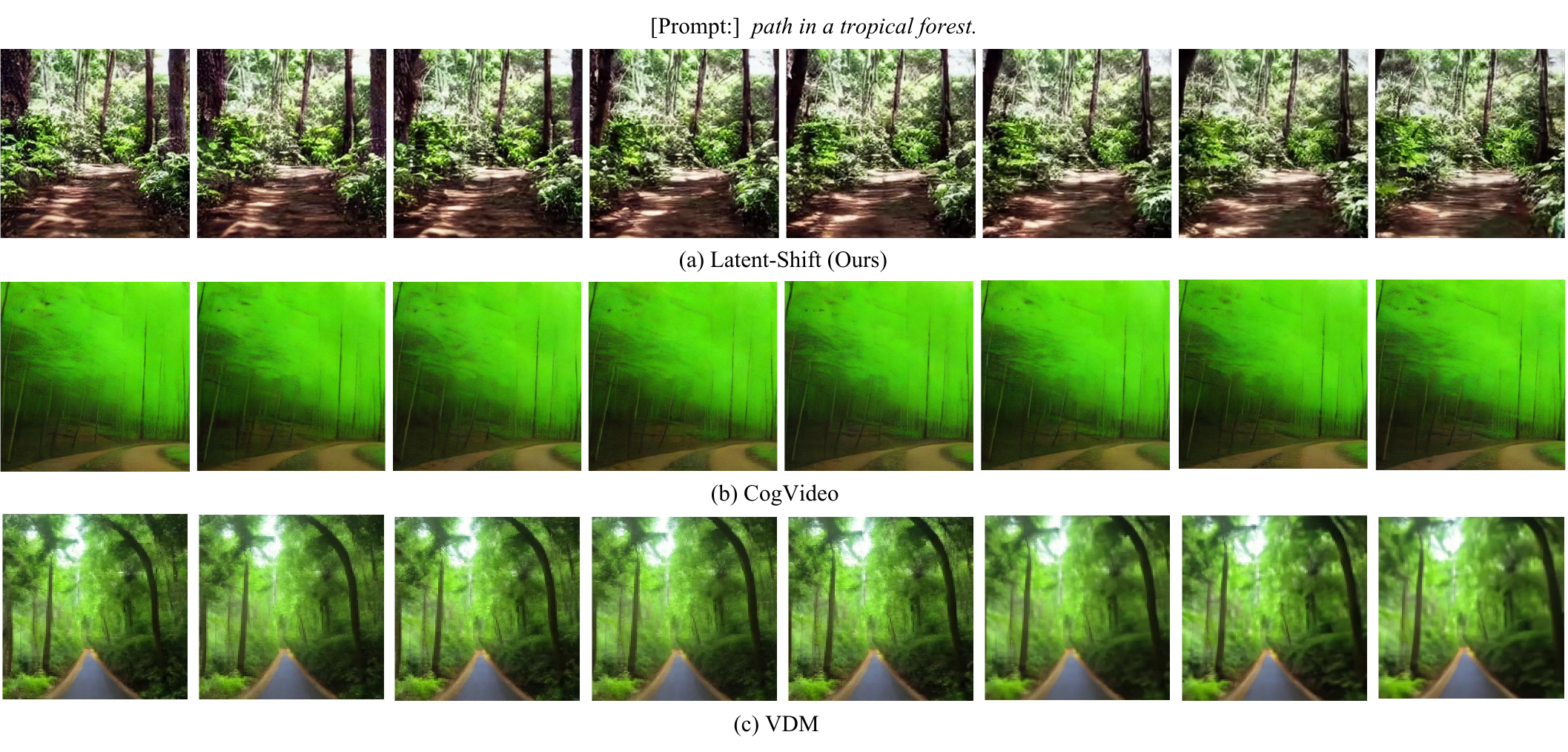}%
\end{subfigure}\hfill
\begin{subfigure}{\linewidth}
\includegraphics[width=\linewidth]{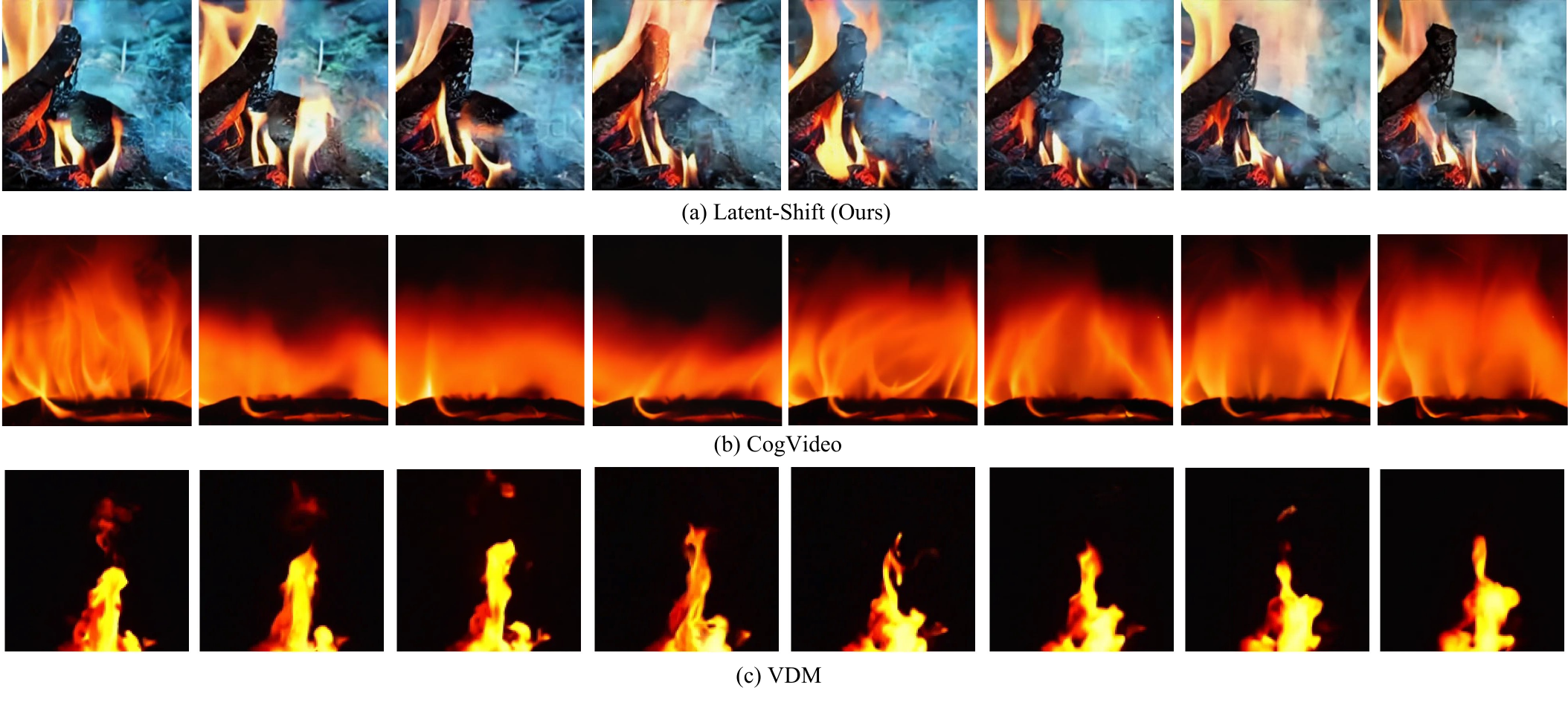}%
\end{subfigure}%
\caption{Text-to-Video generation comparison with CogVideo and Video Diffusion Models (VDM) - continued.}
\label{fig:supp_t2v}
\end{figure*}

\clearpage

\end{document}